# Soil Data Analysis Using Classification Techniques and Soil Attribute Prediction

Jay Gholap, Anurag Ingole, Jayesh Gohil, Shailesh Gargade, Vahida Attar

Dept. of Computer Engineering and IT, College of Engineering,
Pune, Maharashtra-411005, India

**Abstract**

Agricultural research has been profited by technical advances such as automation, data mining. Today ,data mining is used in a vast areas and many off-the-shelf data mining system products and domain specific data mining application soft wares are available, but data mining in agricultural soil datasets is a relatively a young research field. The large amounts of data that are nowadays virtually harvested along with the crops have to be analyzed and should be used to their full extent.
    This research aims at analysis of soil dataset using data mining techniques. It focuses on classification of soil using various algorithms available. Another important purpose is to predict untested attributes using regression technique, and implementation of automated soil sample classification.

## 1. Introduction

Data Mining is a very crucial research domain in recent research world. The techniques are useful to elicit significant and utilizable knowledge which can be perceived by many individuals. Data mining programs consists of diverse methodologies which are predominantly produced and used by commercial enterprises and biomedical researchers. These techniques are well disposed towards their respective knowledge domain. The use of standard statistical analysis techniques is both time consuming and expensive. Efficient techniques can be developed and tailored for solving complex soil data sets using data mining to improve the effectiveness and accuracy of the Classification of large soil data sets [1].
    A soil test is the analysis of a soil sample to determine nutrient content, composition and other characteristics. Tests are usually performed to measure fertility and indicate deficiencies that need to be remedied [2]. The soil testing laboratories are provided with suitable technical literature on various aspects of soil testing, including testing methods and formulations of fertilizer recommendations [4]. It helps farmers to decide the extent of fertilizer and farm yard manure to be applied at various stages of the growth cycle of the crop.
    In a research carried out by Leisa J. Armstrong, comparative study of current data mining techniques such as cluster analysis and statistical methods was carried out to establish the most effective technique. They used a large data set extracted from the Western Australia Department of Agriculture and Food (AGRIC) soils database to conduct this research. The experiments analyzed a small number of traits contained within the dataset to determine their effectiveness when compared with standard statistical techniques [3].
    In our approach, we have developed an automated system for soil classification based on fertility . After obtaining the fertility class labels with the help of automated system, we carried out a comparative study of various classification techniques with the help of data mining tool known as WEKA. The dataset used, was collected from one of the soil testing laboratories in Pune District (Maharashtra, India).Rest of this paper focuses on the prediction of untested attributes. This research has implemented a very sound practical application of linear regression technique by forecasting an obscure property of the soil test.
    The outcome of this research will result into substantial diminution in the price of these tests, which will save a lot of efforts and time of Indian soil testing laboratories.

## 2. Research Methodology

### 2.1. Dataset Collection

The dataset is part of surveys which are carried out regularly in Pune District. Primary data for the soil survey are acquired by field sampling. These samples are then sent for chemical and physical analysis at the soil testing laboratories; hence this dataset was collected from a private soil testing lab in Pune. It contains information about number of soil samples taken from 3 regions of Pune district (Khed, Bhor, and Velhe). Dataset has 9 attributes and a total 1988 instances of soil samples. Table1 describes data collected for each soil sample.

Table 1 : Attribute Description

| Field | Description |
|---|---|
| Ph | pH value of soil |
| EC | Electrical conductivity, decisiemen per meter |
| OC | Organic Carbon, % |
| P | Phosphorous, ppm |
| K | Potassium, ppm |
| Fe | Iron, ppm |
| Zn | Zinc, ppm |
| Mn | Manganese, ppm |
| Cu | Copper, ppm |

### 2.2. Automated System

Soil classification system is essential for the identification of soil properties. Expert system can be a very powerful tool in identifying soils quickly and accurately .Traditional classification systems include use of tables, flow-charts. This type of manual approach takes a lot of time, hence quick, reliable automated system for soil classification is needed to make better utilization of technician's time [9].

We propose an automated system that has been developed for classifying soils based on fertility. Being rule-based system, it depends on facts, concepts, theories which are required for the implementation of this system. Rules for soil classification were collected from soil testing lab. The soil sample instances were classified into the fertility class labels as: Very High, High, Moderately High, Moderate, Low, and Very Low. These class labels for soil samples were obtained with the help of this system and they have been used further for comparative study of classification algorithms.

## 3. A Comparative Study Of Soil Classification

The classification of soil was considered critical to study because depending upon the fertility class of the soil the domain knowledge experts determines which crops should be taken on that particular soil and which fertilizers should be used for the same.
The following section describes Naive Bayes, J48, JRip algorithms briefly.

### 3.1. Naive Bayes

A naive Bayes classifier is a simple probabilistic classifier based on applying Bayes' theorem with strong (naive) independence assumptions. Depending on the precise nature of the probability model, naive Bayes classifiers can be trained very efficiently in a supervised learning setting. An advantage of the naive Bayes classifier is that it only requires a small amount of training data to estimate the parameters (means and variances of the variables) necessary for classification [5].

### 3.2. J48 (C4.5)

J48 is an open source Java implementation of the C4.5 algorithm in the Weka data mining tool. C4.5 is a program that creates a decision tree based on a set of labeled input data. This decision tree can then be tested against unseen labeled test data to quantify how well it generalizes. This algorithm was developed by Ross Quinlan. It is an extension of Quinlan's earlier ID3 algorithm. C4.5 uses ID3 algorithm that accounts for continuous attribute value ranges, pruning of decision trees, rule derivation, and so on.

The decision trees generated by C4.5 can be used for classification, and for this reason, C4.5 is often referred to as a statistical classifier [6].

### 3.3. JRip

This algorithm implements a propositional rule learner, Repeated Incremental Pruning to Produce Error Reduction (RIPPER), which was proposed by William W. Cohen as an optimized version of IREP.

In this paper, three classification techniques (naïve Bayes, J48 (C4.5) and JRip) in data mining were evaluated and compared on basis of time, accuracy, Error Rate, True Positive Rate and False Positive Rate. Tenfold cross-validation was used in the experiment. Our studies showed that J48 (C4.5) model turned out to be the best classifier for soil samples.

Table 2: Comparison of different classifiers

| Classifier | Naïve Bayes | JRip | J48 |
|---|---|---|---|
| Correctly Classified Instances | 765 | 1794 | 1827 |
| Incorrectly Classified Instances | 1223 | 194 | 161 |
| Accuracy | 38.40% | 90.24% | 91.90% |
| Mean Absolute Error | 0.229 | 0.0411 | 0.0299 |

## 4. Prediction Of Untested Attributes

Using regression algorithms like Linear Regression, Least Median Square, Simple Regression different attributes were predicted. According to these results the values of Phosphorous attribute was found to be most accurately predicted and it depends on least number of attributes.

When all attributes are numeric, linear regression is a natural and simple technique to consider for numeric prediction, but it suffers from disadvantage of linearity. If data exhibits non-linear dependency, it may not give good results .In this case, least median square technique is used. Median regression techniques incur high computational cost which often makes them infeasible for practical problems [8]. Several regression tests were carried out using WEKA data mining tool to predict untested numeric attributes. Linear-Regression test for predicting phosphor gave the best and accurate results. These predictions can be used to find out phosphor content without taking traditional chemical tests in soil testing labs, and this will eventually save a lot of time. Statistical results of these tests are given in Table3.

There were very limited variations amongst the predicted values of phosphor attribute. Though the Least Median of Squares algorithms is known to produce better results, we noticed that the accuracy of linear regression was relatively equivalent to that of least median of squares algorithm.

Table 3: Comparisons of Regression Algorithms

| Algorithm | Linear Regression | Least Median Square Regression |
|---|---|---|
| Time taken to build the model | 0.16 s | 10.84 s |
| Relative Absolute Error | 10.77% | 10.01% |
| Correlation Coefficient | 0.9810 | 0.9803 |

Table4 shows some of the phosphor predictions by Linear Regression

Table 4: Predictions on test data

| Actual Value Using Soil Testing | Predicted Value Using Linear Regression | Error |
|---|---|---|
| 10.3 | 10.661 | 0.361 |
| 7.7 | 7.431 | -0.269 |
| 4.6 | 4.653 | 0.053 |
| 9.5 | 8.478 | -1.022 |
| 2.9 | 3.035 | 0.135 |
| 5.1 | 4.915 | -0.185 |
| 15.3 | 15.667 | 0.367 |
| 7 | 7.402 | 0.402 |
| 18.4 | 18.743 | 0.343 |
| 4.4 | 4.388 | -0.012 |
| 13.5 | 13.438 | -0.062 |

Here the Relative Absolute Error is nearly same for both the prediction algorithms. Even though Least Median Square regression gives better numeric predictions but the time taken to build the model is 67 times that of Linear Regression, hence computational cost used by Linear Regression is much lower than that of least median square technique.

## 5. Conclusion and Future Work

In this paper, we have proposed an analysis of the soil data using different algorithms and prediction technique. In spite the fact that the least median squares regression is known to produce better results than the classical linear regression technique, from the given set of attributes, the most accurately predicted attribute was "P" (Phosphorous content of the soil) and which was determined using the Linear Regression technique in lesser time as compared to Least Median Squares Regression. We have demonstrated a comparative study of various classification algorithms i.e. Naïve Bayes, J48 (C4.5), JRip with the help of data mining tool WEKA. J48 is very simple classifier to make a decision tree, but it gave the best result in the experiment. In future, we contrive to build Fertilizer Recommendation System which can be utilized effectively by the Soil Testing Laboratories. This System will recommend appropriate fertilizer for the given soil sample and cropping pattern.